\documentclass[letterpaper]{article} 
\usepackage[preprint]{aaai2027} 
\usepackage[hyphens]{url} 
\usepackage{graphicx} 
\usepackage{natbib} 
\usepackage{caption} 
\usepackage{xcolor}
\usepackage{enumitem}
\usepackage{amsmath}
\usepackage{amssymb}
\usepackage{booktabs}

\usepackage[most]{tcolorbox}

\newcommand{\CnCompanyNames}{\raisebox{-0.14em}{\includegraphics[height=0.92em]{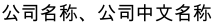}}}
\newcommand{\CnFormerNames}{\raisebox{-0.14em}{\includegraphics[height=0.92em]{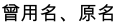}}}
\newcommand{\CnNameChange}{\raisebox{-0.14em}{\includegraphics[height=0.92em]{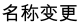}}}
\newcommand{\CnStatementTypes}{\raisebox{-0.14em}{\includegraphics[height=0.92em]{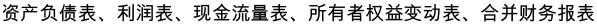}}}
\newcommand{\CnJudgmentLabelSix}{\raisebox{-0.14em}{\includegraphics[height=0.92em]{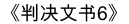}}}
\newcommand{\CnJudgmentLabel}{\raisebox{-0.14em}{\includegraphics[height=0.92em]{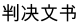}}}

\definecolor{EnSIBlue}{HTML}{163D8C}
\definecolor{EnSITeal}{HTML}{00796B}
\definecolor{EnSIGreen}{HTML}{238636}
\definecolor{EnSIPurple}{HTML}{6F42C1}
\definecolor{EnSIBurgundy}{HTML}{8B1E3F}
\definecolor{EnSIOrange}{HTML}{A65100}

\newtcolorbox{EnSIPromptBox}[2]{
  enhanced,
  breakable,
  title={#1},
  fonttitle=\bfseries,
  coltitle=white,
  colbacktitle=#2,
  colframe=#2,
  colback=#2!4,
  boxrule=0.6pt,
  arc=1.5mm,
  outer arc=1.5mm,
  left=2.2mm,
  right=2.2mm,
  top=1.8mm,
  bottom=1.8mm,
  boxsep=0pt,
  before skip=5pt,
  after skip=9pt
}

\newenvironment{EnSIPromptText}
  {\ttfamily\small\raggedright
   \setlength{\parindent}{0pt}%
   \setlength{\parskip}{0pt}%
   \hyphenchar\font=-1\relax}
  {\par}

\newcommand{\PromptLine}[1]{\noindent#1\par}
\newcommand{\PromptBlank}{\noindent\mbox{}\par}

\title{EnSI-RAG: Entity-Structure-Indexed Retrieval-Augmented Generation for
Long-Document Question Answering}

\author{
Xuanyu Meng,
Jiashuo Sun,
Jash Rajesh Parekh,
Jiawei Han
}

\affiliations{
University of Illinois Urbana-Champaign\\
Urbana, Illinois, USA\\
\{xuanyum2, jiashuo5, jashrp2, hanj\}@illinois.edu
}

\begin{document}

\maketitle

\begin{abstract}
Question answering (QA) over long, connected documents remains
challenging because relevant evidence may span multiple entities and
their relationships. Existing retrieval-augmented generation (RAG)
methods typically index documents as raw chunks and retrieve them
through embedding similarity. Their performance degrades when chunk
boundaries separate entities from supporting evidence or when a
question requires multi-hop reasoning across the corpus. We propose
\textit{EnSI-RAG} (\textbf{En}tity-\textbf{S}tructure-\textbf{I}ndexed
\textbf{R}etrieval-\textbf{A}ugmented \textbf{G}eneration), a framework
that constructs a query-independent, entity-centered index. Each record
$\langle e,t,k,v\rangle$ represents an entity $e$, its type $t$, a
semantic category
$k\in\{\text{property},\text{relation},\text{aspect}\}$, and a value
$v$, while retaining links to the original source passages. At query
time, these records serve as retrieval handles, and an LLM synthesizes
the retrieved passages into the final answer. This design separates
evidence localization from answer synthesis while preserving traceable
source evidence. Across Loong and Oolong, EnSI-RAG achieves an average
accuracy of 78.24. Relative to the published baseline scores used as
references, this is 6.62 points higher, suggesting its effectiveness
across these settings. The code is available at
\textcolor{blue}{\url{https://github.com/RamonMeng/EnSI-RAG}}.
\end{abstract}

\section{Introduction}

Answering questions over long documents remains a fundamental challenge for large language models (LLMs). In many real-world applications, the information required to answer a question is rarely contained within a single passage. Financial analysts may need to compare values across multiple reports, legal researchers may need to classify or aggregate evidence from many judgments, and scientists may need to trace citations, methods, or findings across collections of papers. These tasks require systems not only to locate relevant evidence, but also to preserve entity identity, document structure, and relationships among dispersed pieces of information.

Recent advances in LLMs have made long-document question answering increasingly feasible. One prominent direction is to extend the context window, allowing models to consume longer inputs directly \citep{anthropic2024claude35,geminiteam2024gemini15}. However, larger context windows alone do not guarantee reliable evidence utilization: models remain sensitive to the position of relevant information, exhibiting the well-known ``lost in the middle'' phenomenon \citep{liu2024lost}. Moreover, recent long-context benchmarks demonstrate that even frontier models continue to struggle with questions requiring evidence aggregation and multi-hop reasoning across long documents rather than localized information lookup \citep{wang2024loong,bertsch2025oolong,bai2025longbenchv2}. Another direction is retrieval-augmented generation (RAG), which retrieves external evidence before generation \citep{lewis2020rag,guu2020realm}. While RAG substantially reduces the amount of text processed by the LLM, it shifts the burden to the retriever, since answer quality is ultimately limited by the evidence it retrieves. 

Most existing RAG systems treat documents as collections of fixed-length chunks and retrieve them using lexical or dense similarity. Although simple and scalable, fixed-size chunking is often poorly aligned with the semantic organization of long documents. Chunk boundaries may separate entities from their supporting evidence, detach table rows from their headers, or merge multiple unrelated topics into the same retrieval unit. Consequently, retrieval may return passages that are locally similar to the query but insufficient for answering it, while overlooking semantically essential evidence that shares little lexical similarity with the question.

Our key observation is that retrieval units should be defined semantically rather than mechanically. Instead of indexing arbitrary chunks, EnSI-RAG represents documents as \emph{entity-centered passages}, where each passage focuses primarily on one entity together with the local context describing that entity. Such passages preserve semantically coherent evidence while reducing both semantic fragmentation and contamination, providing a more reliable foundation for retrieval. However, semantically coherent passages alone are insufficient. To retrieve them efficiently for arbitrary future questions, they must also be organized into a query-independent representation that exposes their underlying semantic structure.

Based on this observation, we propose \textbf{EnSI-RAG} (Entity-Structure-Indexed Retrieval-Augmented Generation), a query-independent framework for long-document question answering. EnSI-RAG first preprocesses a document collection by constructing entity-centered passages and extracting structured records describing the entities, their semantic types, and their associated properties, relations, and aspects. These records are organized into an entity-structure index that serves as a set of retrieval handles pointing back to the original passages. At query time, the index guides retrieval toward the supporting passages, while the LLM remains responsible for integrating the retrieved evidence and synthesizing the final answer. Unlike fully structured approaches that perform reasoning over normalized databases, EnSI-RAG uses structure only to improve evidence localization; reasoning remains grounded in the original document text.

We evaluate EnSI-RAG on two complementary long-document
question-answering benchmarks: Loong \citep{wang2024loong} and
Oolong \citep{bertsch2025oolong}. Loong evaluates heterogeneous
multi-document reasoning across financial, legal, and academic
domains, while Oolong emphasizes large-scale information aggregation
over long inputs. Across these different document structures and
reasoning requirements, EnSI-RAG applies a unified retrieval framework
while allowing domain-specific passage construction. Experimental
results demonstrate that this design achieves competitive or superior
performance compared with strong retrieval baselines.
Our contributions are threefold. First, we introduce an entity-centered passage representation that defines retrieval units according to semantic coherence rather than fixed-length segmentation. Second, we propose a query-independent entity-structure index that enables passages to be retrieved through entities and their associated semantic information while preserving the original documents as evidence. Third, we develop a unified retrieval framework that combines structured evidence localization with LLM-based semantic synthesis, providing an effective middle ground between conventional chunk-based RAG and fully structured question answering systems.

\section{Related Work}

\noindent\textbf{Offline corpus structuring.}
Building query-independent corpus representations of corpora long predates LLMs. Classical information retrieval constructs inverted indexes and term-weighted vectors offline \citep{salton1988term, robertson1994okapi}, while web-scale information extraction extends indexing to entities and relations \citep{etzioni2004web, etzioni2008open, dong2014knowledge}. Recent methods use LLMs to construct hierarchical or recursive summaries \citep{sarthi2024raptor, chen2023walking}, segment text into propositions or entity-centered spans \citep{chen2024dense, jiang2024longrag}, or convert documents into tables, schemas, and knowledge graphs \citep{li2024structrag, litecost2026, wu2022text2table}. Such representations provide compact and uniform access to a corpus, but information discarded during structuring is unavailable at query time, and schemas designed for one question class may transfer poorly to others. EnSI-RAG instead uses structured records only as retrieval handles: they locate the original passages, which retain the evidence used for answer generation.

\vspace{1em}
\noindent\textbf{Retrieval-augmented generation.}
Larger LLM context windows do not fully resolve long-document QA as context size is not the binding constraint \citep{du2025context}. Models exhibit a lost in the middle effect, retrieving evidence unevenly across input positions \citep{liu2024lost}, leading to struggles  with aggregation-intensive questions \citep{wang2024loong, bertsch2025oolong, bai2025longbenchv2, yen2025helmet}. RAG grounds generation in retrieved evidence \citep{lewis2020rag, guu2020realm, izacard2023atlas}, with adaptive, iterative, and RL-trained retrieval policies improving this process \citep{asai2024selfrag, trivedi2023ircot, jiang2023flare, jin2025searchr1}. However, these methods still rely on near-perfect retriever recall. The dominant approach, ranking fixed-size chunks by lexical or dense similarity \citep{karpukhin2020dpr, izacard2022contriever, khattab2020colbert}, is poorly suited to long documents. For example, chunk boundaries may sever coherent discussions, detach table rows from their headers, or merge unrelated entities into a single retrieval unit, yielding context that is locally similar yet incomplete or omitting evidence with little lexical overlap \citep{gong2020recurrent, wang2025document, shin2025multidocfusion}. Graph-based RAG addresses part of this issue by building entity and relation structures \citep{edge2024graphrag, gutierrez2024hipporag, parekh2025structure, guo2024lightrag}, but is primarily developed and evaluated on short-passage benchmarks with well-canonicalized entities. These assumptions often break down in long, heterogeneous, and domain-specific documents.

\section{EnSI-RAG}

\begin{figure*}[t]
\centering
\includegraphics[width=\textwidth]{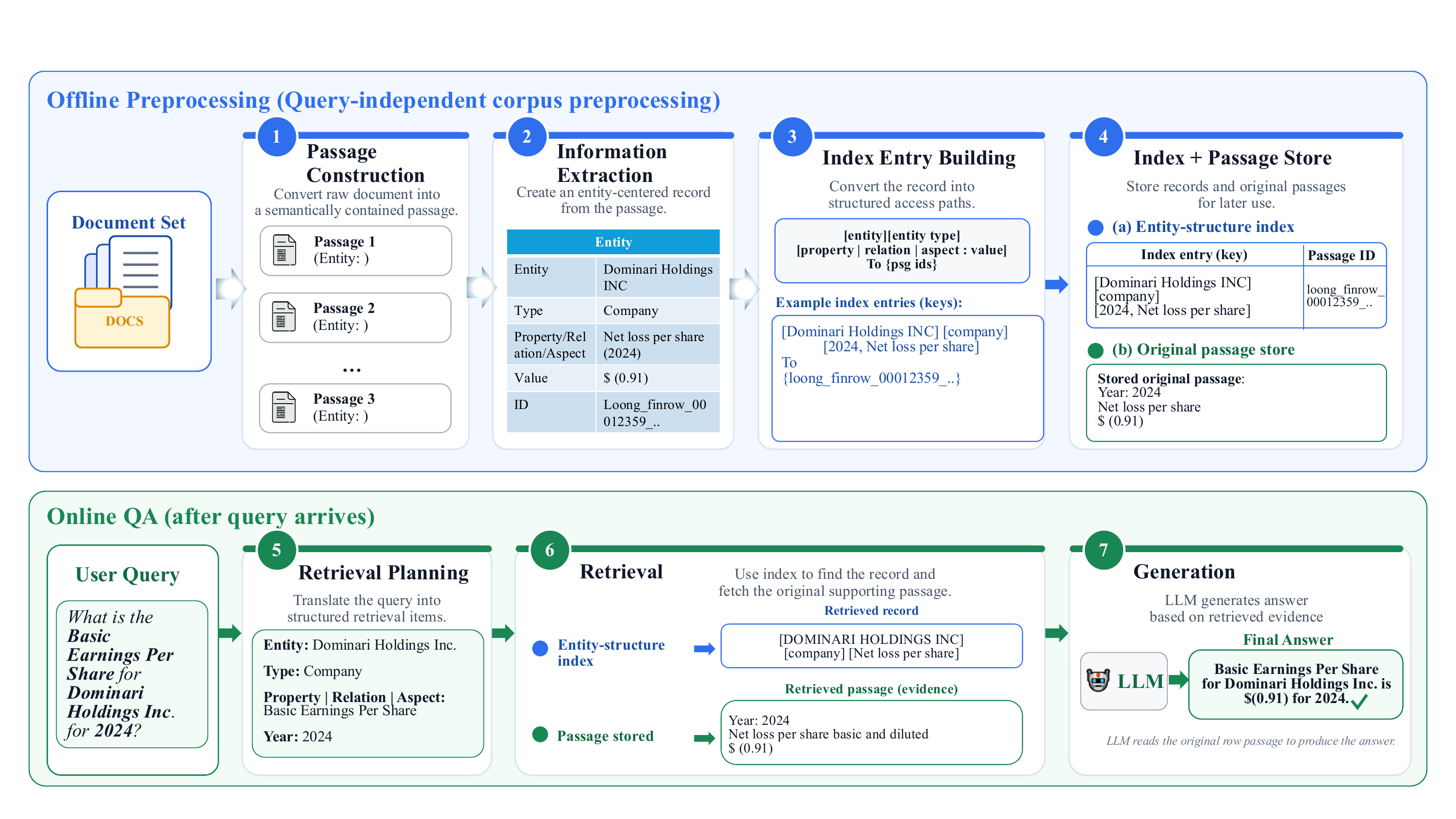}
\caption{End-to-end EnSI-RAG workflow on a financial QA example. The upper pipeline shows query-independent offline
preprocessing: a Dominari Holdings filing is converted into a self-contained financial row passage, an entity-centered record,
and entity-structure index entries. The lower pipeline shows online QA: after the query arrives, EnSI-RAG plans retrieval,
retrieves the matched record and original supporting passage, and uses an LLM to generate the final answer from the passage
evidence}
\label{fig:ensi-overview}
\end{figure*}

Figure~\ref{fig:ensi-overview} presents the overall workflow of EnSI-RAG. The framework consists of an offline preprocessing stage followed by an online question-answering stage. During offline preprocessing, EnSI-RAG performs \textbf{Passage Construction} to convert long documents into semantically contained, entity-centered passages, followed by \textbf{Information Extraction} to derive structured records describing the entities appearing in each passage together with their semantic types, properties, relations, and aspects. These records are then transformed during \textbf{Index Entry Building} into a query-independent entity-structure index, while the original passages are preserved in a passage store. During online question answering, the \textbf{Retrieval} stage first analyzes the user query to derive structured retrieval requirements and then uses the entity-structure index to locate the corresponding supporting passages. Finally, the retrieved original passages are provided to the \textbf{Generation} stage, where an LLM synthesizes the final answer directly from grounded source evidence. This separation between offline indexing and online reasoning allows EnSI-RAG to reuse the same indexed corpus for arbitrary future queries while preserving the original passages as the basis for answer generation. The following subsections describe these stages in detail.

\subsection{Passage Construction}

Given a collection of long documents $\mathcal{D}=\{d_1,\ldots,d_n\}$, EnSI-RAG converts each document into semantically self-contained passages. Unlike a fixed-size chunk, which is determined by a token budget, character length, or sliding window, a passage is organized around a main entity and retains sufficient local context to identify its type and extract its properties, relations, and aspects. Formally,
\begin{equation}
\begin{aligned}
\mathcal{P}_d &= \{p_1,\ldots,p_m\},\\
\mathcal{P} &= \bigcup_{d\in\mathcal{D}}\mathcal{P}_d.
\end{aligned}
\end{equation}
Each passage has a stable identifier $\textit{psg\_id}$ and preserves its source document, available source offsets, and lightweight structural context, such as titles, headers, reporting periods, or local citations.

Passage construction seeks semantic containment: a passage should be
able to keep an entity with its supporting evidence, yet focused
enough to avoid mixing unrelated entities or facts. Rather than imposing
a globally fixed passage schema, EnSI-RAG instantiates passage design
on the fly during corpus preprocessing according to the requirements of
the task and the structural organization of the documents. Passage
boundaries may therefore vary across tasks, domains, and document types,
while remaining independent of any individual query. Regardless of the
constructor used, every passage conforms to the same downstream
interface---a semantically contained text unit centered on one entity and
linked to its source by a unique $\textit{psg\_id}$.

This objective balances evidence coverage for the main entity against the inclusion of unrelated entities or topics. These are semantic criteria rather than hard length constraints, so passage sizes may vary with the context needed to make the evidence self-contained.

Compared with arbitrary chunking, this design reduces semantic fragmentation because an entity and its evidence remain together whenever possible. It also reduces semantic contamination because unrelated facts are less likely to share a retrieval unit. The resulting context makes the extraction of entity types, properties, relations, and aspects more reliable. Most importantly, each passage preserves the original text and its provenance, allowing later stages to retrieve grounded evidence rather than an extraction-only surrogate and to defer final semantic integration to the LLM.

\subsection{Information Extraction}

EnSI-RAG next extracts entity-centered information from each passage. Rather than answering a particular question or normalizing the corpus into a rigid database, this stage creates reusable retrieval handles through which passages can be addressed by their entity structure.

For each $p\in\mathcal{P}$, the extractor produces
\begin{equation}
\begin{aligned}
\mathcal{R}_p &= \{r_1,\ldots,r_k\},\\
r &= \langle e,t,\mathcal{M},\textit{psg\_id}\rangle,
\end{aligned}
\end{equation}
where $e$ is an entity mention or canonical name, $t$ is its type, and $\mathcal{M}$ contains its passage-supported semantic fields. Each field is
\begin{equation}
\begin{aligned}
m &= \langle c,k,v\rangle,\\
c &\in \{\textit{property},\textit{relation},\textit{aspect}\},
\end{aligned}
\end{equation}
where $k$ is the field name and $v$ its value. A property records an attribute value, a relation records a target entity, and an aspect indicates the kind of information expressed about the entity. For example, $\textit{birth-loc}=\textit{Memphis}$ is a property of a singer, whereas $\textit{cites}=\textit{paper B}$ is a relation of \textit{paper A}. Birthplace and financial metrics are properties, citations and affiliations are relations, and method or legal reasoning are aspects.

These categories describe distinct retrieval cues. Properties connect an entity to an attribute value, relations connect it to another entity, and aspects expose passages that discuss a particular facet even when no single normalized value is appropriate. A passage may yield multiple records and each record may contain multiple supported fields, while the common $\textit{psg\_id}$ preserves their connection to the same source evidence.

Extraction follows two principles. First, every indexed field must be supported by local passage evidence. Second, extraction is retrieval-oriented rather than answer-oriented: a record need not encode every detail required by future questions, but should provide reliable access to the original evidence. This differs from fully structured systems, in which an extraction error directly enters the reasoning state. EnSI-RAG uses records primarily as passage-addressing keys and generates answers from the retrieved source passages; consequently, even a partial record remains useful when it leads to the correct evidence.

This separation is essential: extracted values identify where information is located but are not treated as its complete representation. Details unnecessary for indexing remain in the passage and are available during generation. Conversely, unsupported information is excluded even if it appears plausible from background knowledge, keeping every access path grounded in the corpus.

\subsection{Index Entry Building}

The extracted records form an entity-structure index $\mathcal{I}$ that maps each tuple $(e,t,c,k,v)$ to the identifiers of passages whose records contain the corresponding entity, type, and semantic field. An entry is written compactly as
\begin{equation}
[e][t][c:k=v]\rightarrow\{\textit{psg\_id}\}.
\end{equation}
Every field in a record contributes its $\textit{psg\_id}$ to the corresponding set. This rule covers all three field categories and naturally aggregates evidence across passages and documents. For instance,
\[
\begin{aligned}
&[\textit{Franklin}][\textit{Singer/Person}]\\
&[\textit{property}:\textit{birth-loc}=\textit{Memphis}]\\
&\hspace{2em}\rightarrow\{\textit{psg\_id}\}.
\end{aligned}
\]
The output is deliberately a set rather than a single identifier. The same entity-structure key may be supported by passages in which a fact is repeated, refined, contradicted, or presented from different perspectives. Retaining all such passages avoids prematurely collapsing evidence into one canonical passage or normalized value.

Relations that are useful in both directions may also receive inverse entries. For example, a \textit{paper A cites paper B} entry may be accompanied by an entry from \textit{paper B} to \textit{paper A} under \textit{cited-by}. The same principle applies to authorship, aliases, location containment, and organizational affiliation, enabling retrieval from either side of a relation.

Set-valued indexing supports different reasoning needs. A narrow lookup can select the most relevant passage; aggregation or comparison can combine several passages; and ambiguous or multi-hop questions can explore alternative evidence paths. The index is therefore a passage-addressing layer rather than a lossy database of final answers.

The retained alternatives matter because repeated evidence is not necessarily redundant. Separate passages may supply complementary context, use different terminology, qualify a value, or reveal a disagreement that should be resolved during generation. Keeping these passages available lets retrieval select evidence according to the current question rather than fixing a single preferred context during indexing.

Each entry retains both symbolic fields and a textual representation. The former enables constrained matching over entities, types, categories, field names, and values, while the latter supports semantic matching across different surface forms. Together they provide precise yet flexible access to the source corpus.

\subsection{Retrieval}

Given a question $q$, EnSI-RAG derives a retrieval plan
\begin{equation}
H(q)=\{h_1,\ldots,h_m\},
\end{equation}
where each hop $h_i$ specifies a retrieval requirement over some combination of entity, entity type, property, relation, aspect, value, or target entity. Direct questions may require only one hop; in multi-hop questions, later requirements may depend on entities or values discovered earlier.

Accordingly, $H(q)$ is a plan rather than a requirement to decompose every question. When one structured requirement is sufficient, EnSI-RAG performs a single retrieval step. When dependencies exist, the result of hop $h_i$ supplies an intermediate entity or value used to instantiate $h_{i+1}$. This preserves the original question's reasoning chain while expressing each retrieval step in the same entity-centered vocabulary as the index.

Each hop is matched against an index key $x=(e,t,c,k,v)$. Their compatibility is modeled as
\begin{equation}
\operatorname{score}(h,x)
=\sum_{a\in\{e,t,c,k,v\}}\lambda_a s_a(h,x),
\end{equation}
where $a$ ranges over entity, entity type, field category, field name, and value or target; $s_a$ measures compatibility on field $a$; and $\lambda_a$ controls its contribution. This formulation accommodates both symbolic and semantic matching without tying the framework to a particular implementation.

The component form makes partial requirements possible. A hop may strongly constrain an entity and relation while leaving the target unknown, or it may seek an aspect for an entity of a particular type. Only fields expressed or implied by the retrieval requirement need to contribute to its match. The ranked keys therefore act as structured hypotheses about which index entries provide the required evidence.

Because every key maps to a passage set, the candidates for a hop and the accumulated evidence for the question are
\begin{equation}
\begin{aligned}
\mathcal{C}_h
&=\bigcup_{x\in\operatorname{TopKeys}(h)}\mathcal{I}(x),\\
\mathcal{C}_q
&=\bigcup_{h\in H(q)}\mathcal{C}_h.
\end{aligned}
\end{equation}
Duplicate identifiers are removed, after which the passages may be reranked before generation.

Key ranking and passage ranking serve different roles. The former identifies compatible entity structures; the latter selects the most useful source contexts among all passages reached through those structures. Their separation is particularly useful when a highly compatible key has several supporting passages of different relevance to the question.

Unlike ordinary RAG, EnSI-RAG does not rely solely on matching the full question against anonymous raw chunks. It uses structured, entity-centered keys to constrain retrieval while returning the original evidence passages. In multi-hop reasoning, one hop can retrieve a relation and expose an intermediate entity that becomes the query entity of the next hop. Retrieval thus follows the corpus's entity structure rather than a single embedding search.

The set-valued index preserves alternative supporting contexts within this structured search space. It can supply multiple passages under one key, compare evidence associated with different keys, and maintain competing paths for ambiguous questions. EnSI-RAG therefore indexes structured access paths while leaving evidence integration and final answer generation to the LLM.

\subsection{Generation}

After retrieval, EnSI-RAG generates the final answer directly from the retrieved original passages. Given a user question $q$ and the retrieved passage set $\mathcal{C}_q$, the generation stage simply provides them to an LLM:

\begin{equation}
\hat{y}=\mathrm{LLM}(q,\mathcal{C}_q).
\end{equation}

Unlike the preceding stages, this step performs no additional indexing or symbolic reasoning. The LLM is responsible only for synthesizing the final answer from the retrieved evidence. Because retrieval has already localized the relevant supporting passages, the generation model can focus on semantic understanding and evidence integration rather than searching through the document collection.

This design cleanly separates evidence localization from answer synthesis. EnSI-RAG uses the entity-structure index to retrieve the original supporting passages, while the LLM operates directly on those passages to produce the final answer. Consequently, the retrieved passages remain the source of evidence, and the structured records are used only as retrieval handles rather than as a reasoning substrate.

\section{Experiments}

\subsection{Experimental Setup}

We evaluate EnSI-RAG on two open-source long-document question-answering benchmarks: Loong~\cite{wang2024loong} and Oolong~\cite{bertsch2025oolong}. Loong emphasizes heterogeneous multi-document reasoning across financial, legal, and academic domains, whereas Oolong focuses on large-scale information aggregation over long inputs. Together, they test whether EnSI-RAG can localize and integrate evidence across different document structures and reasoning requirements. The benchmark details are shown in Table~\ref{tab:benchmarks}.

We follow the official evaluation protocol of each benchmark. Loong is evaluated using average accuracy. For Oolong, non-numeric questions use an LLM-based judge, whereas numeric aggregation questions use the official deviation-based metric.

We compare EnSI-RAG with RAG~\cite{lewis2020rag},
LongRAG~\cite{jiang2024longrag},
GraphRAG~\cite{edge2024graphrag}, BaseModel,
DocETL~\cite{shankar2024docetl},
Chain-of-Agent~\cite{zhang2024chainofagents},
RLM~\cite{zhang2025rlm}, and
SLIDERS~\cite{joshi2026sliders}. All baseline scores on
Loong and Oolong are taken from SLIDERS~\cite{joshi2026sliders}.
EnSI-RAG is evaluated on our sampled Loong subset and on the Oolong
subset obtained by filtering for the required document length and
selecting the first 192 eligible questions in dataset order, following
the SLIDERS configuration. Because our Loong evaluation subset may
differ from that used for the published baselines, its baseline scores
are included as reference rather than as controlled head-to-head
comparisons.
\begin{table}[t]
\centering
\small
\setlength{\tabcolsep}{4pt}
\begin{tabular}{lcc}
\toprule
\textbf{Benchmark} & \textbf{Task Type} & \textbf{\# Questions} \\
Loong        & Retrieval, Aggregation      & 177 \\
Oolong       & Classification, Aggregation & 192 \\
\bottomrule
\end{tabular}
\caption{Benchmarks used in our experiments. The number of questions corresponds to the evaluated subset of each benchmark.}
\label{tab:benchmarks}

\end{table}

\subsection{Main Results}

Table~\ref{tab:main_results} compares EnSI-RAG with the baseline
results reported by SLIDERS~\cite{joshi2026sliders}. The Oolong
results use the same deterministic benchmark slice, whereas the Loong
results may use different sampled subsets and are therefore included
as reference rather than as a controlled head-to-head comparison.
Each EnSI-RAG result is obtained from one complete run.
\begin{table*}[t]
\centering
\begin{tabular}{l l c c c}
\toprule
\textbf{Models} & \textbf{LLMs}
& \textbf{Oolong}
& \textbf{Loong}
& \textbf{Avg.} \\
\midrule
RAG
& Qwen3-4B + GPT-4.1
& 11.32 & 54.35 & 32.84 \\

LongRAG
& Qwen3-4B + GPT-4.1
& 22.00 & 59.10 & 40.55 \\

GraphRAG
& Qwen3-4B + GPT-4.1
& 22.00 & 61.28 & 41.64 \\
\midrule

BaseModel
& GPT-4.1
& 45.56 & 76.74 & 61.15 \\

BaseModel
& Qwen3.5-122B-A10B
& 24.89 & 74.78 & 49.84 \\
\midrule

DocETL
& GPT-4.1
& 49.00 & 75.03 & 62.02 \\

Chain-of-Agent
& GPT-5 + GPT-5-mini
& 17.11 & 54.46 & 35.79 \\

RLM
& GPT-5 + GPT-5-mini
& 51.42 & 72.64 & 62.03 \\
\midrule

SLIDERS
& GPT-4.1 + GPT-4.1-mini
& 64.67 & 78.57 & 71.62 \\
\midrule

\textbf{EnSI-RAG}
& GPT-4.1 + GPT-4.1-mini
& \textbf{71.84} & \textbf{84.64} & \textbf{78.24} \\
\bottomrule
\end{tabular}

\caption{Accuracy on the two benchmarks. All baseline scores are
taken from SLIDERS~\cite{joshi2026sliders}, while EnSI-RAG results are
obtained from our experiments. Higher is better.}
\label{tab:main_results}
\end{table*}

\begin{table*}[t]
\centering
\small
\setlength{\tabcolsep}{8pt}
\begin{tabular}{lccc}
\toprule
\textbf{Pipeline Stage} &
\textbf{Input Tokens} &
\textbf{Latency (s)} &
\textbf{LLM Calls} \\
\midrule
\multicolumn{4}{l}{\textit{Offline preprocessing}} \\
Passage Construction and Information Extraction      & 157,031 & 468.51 & 32.35 \\
Index Entry Building      & 0       & 0.05   & 0.00  \\
\midrule
\multicolumn{4}{l}{\textit{Online question answering}} \\
Retrieval                 & 656     & 19.01  & 1.00  \\
Generation                & 5,460   & 6.73   & 1.00  \\
\midrule
\textbf{Overall}          & \textbf{163,146} & \textbf{494.30} & \textbf{34.35} \\
\bottomrule
\end{tabular}
\caption{Average efficiency per query under the default EnSI-RAG configuration. Passage Construction, Information Extraction, and Index Entry Building are offline stages whose costs are amortized over the evaluated queries. Retrieval and Generation are online stages. In the current implementation, LLM-based information extraction is integrated into Passage Construction.}
\label{tab:efficiency}
\end{table*}

\begin{figure*}[!t]
\centering
\includegraphics[width=\textwidth]
{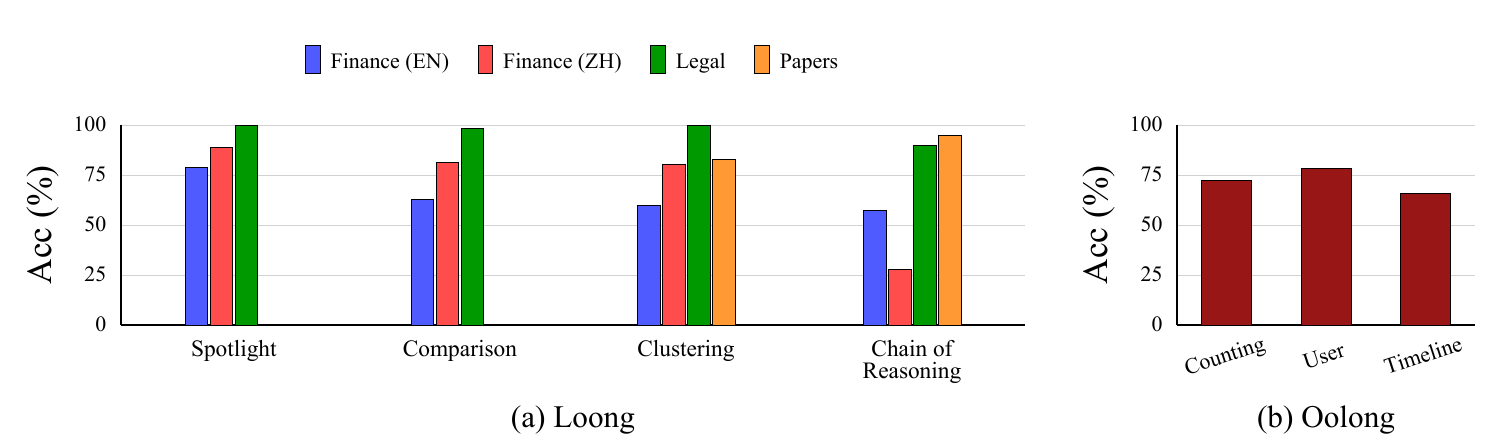}
\caption{Accuracy breakdown by task type on Loong and Oolong.
Missing bars denote absent domain--task combinations. Higher is better.}
\label{fig:benchmark-breakdown}
\end{figure*}

\subsection{Ablation Study}
\label{sec:ablation}

We study three design choices in EnSI-RAG: passage granularity,
property/aspect granularity, and retrieval depth. For passage
granularity and retrieval depth, we report results on Financial
questions from Loong. We evaluate property/aspect granularity on
Loong questions from the Financial, Paper, and Legal domains. Within
each comparison, we use the same questions and vary only the factor
under study. Following the evaluation protocol of Loong
\cite{wang2024loong}, an evaluator assesses each answer from 1 to 100
according to accuracy, hallucination, and completeness. We report
average accuracy (Acc.).

\begin{table}[t]
\centering
\setlength{\tabcolsep}{12pt}
\begin{tabular}{lc}
\toprule
Passage granularity & Accuracy (\%) $\uparrow$ \\
Row-level   & \textbf{100.00} \\
Table-level & 91.50 \\
\bottomrule
\end{tabular}
\caption{Passage-granularity ablation on English Financial questions from
Loong. Higher is better.}
\label{tab:ablation-passage}
\end{table}

\paragraph{Passage granularity.}
Representing each financial table row as an individual passage
substantially outperforms representing the entire table as one passage.
Row-level passages achieve 100.00, compared with 91.50 for table-level
passages. The 8.50-point gap indicates that compact evidence units
reduce irrelevant context within a retrieved passage and provide a
more precise interface between retrieval and generation. We therefore
select row-level passages for Financial documents.

\begin{figure}[t]
\centering
\includegraphics[width=\columnwidth]{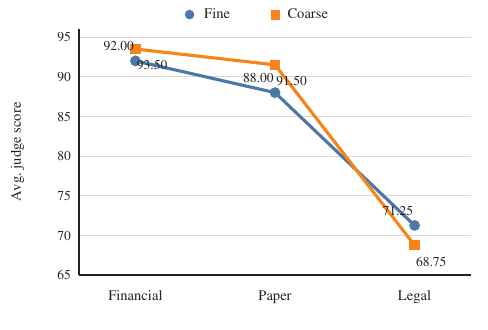}
\caption{Accuracy under different property, relation, and aspect granularities across Loong
domains. Coarse labels perform better on Financial and Paper, whereas
fine-grained labels perform better on Legal. Higher accuracy is better.}
\label{fig:ablation-granularity}
\end{figure}

\paragraph{Property, relation, and aspect granularity.}
We compare labels extracted verbatim from the source with labels mapped
to higher-level concepts. This transformation applies to property and
aspect names; directional relation labels are preserved to avoid
changing graph semantics. Coarse labels improve English Financial from 92.00
to 93.50 and Paper from 88.00 to 91.50, suggesting that semantic
abstraction reduces lexical sparsity in these domains. In contrast,
Legal decreases from 71.25 to 68.75. Legal queries often depend on
fine distinctions among case attributes, and collapsing these
distinctions can make retrieved evidence less discriminative. Thus,
the optimal granularity is domain dependent: coarse labels are selected
for English Financial and Paper, whereas fine labels are retained for Legal.

\begin{figure}[t]
\centering
\includegraphics[width=\columnwidth]{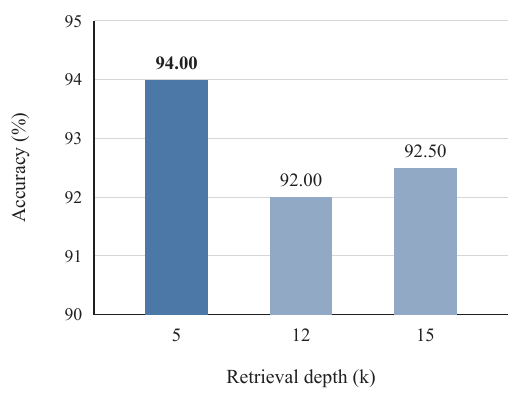}
\caption{Retrieval-depth ablation on English Financial questions
from Loong. Top-5 achieves the highest accuracy; retrieving
additional passages provides no improvement in this setting.
Higher accuracy is better.}
\label{fig:ablation-depth}
\end{figure}

\paragraph{Retrieval depth.}
Top-5 achieves the highest English Financial accuracy of 94.00,
outperforming Top-15 at 92.50 and Top-12 at 92.00. Retrieving more
evidence therefore provides no benefit on this subset, suggesting that
concentrated evidence reduces distraction. Nevertheless, all main
results use Top-12 to maintain a common configuration and avoid
benchmark-specific tuning.

\paragraph{Discussion.}
Overall, the ablations favor row-level passages and domain-dependent
property granularity, while showing that shallower retrieval may
benefit the evaluated English Financial subset.

\subsection{Efficiency}

We evaluate the average per-query efficiency of each EnSI-RAG stage using LLM input tokens, wall-clock latency, and the number of LLM calls. Passage Construction, Information Extraction, and Index Entry Building are query-independent offline stages, so their costs are amortized over the evaluated queries. In the current implementation, passage construction and LLM-based information extraction are performed jointly.

As shown in Table~\ref{tab:efficiency}, Passage Construction and Information Extraction account for 157,031 input tokens, 468.51 seconds, and 32.35 LLM calls per query after amortization. Index Entry Building adds no input tokens or LLM calls and requires 0.05 seconds. The online stages are substantially lighter: Retrieval uses 656 input tokens, 19.01 seconds, and 1.00 LLM call, while Generation uses 5,460 input tokens, 6.73 seconds, and 1.00 LLM call. Overall, EnSI-RAG uses 163,146 input tokens, 494.30 seconds, and 34.35 LLM calls per query. This profile reflects the system's central design: expensive corpus processing is performed offline and reused, while online inference is limited to structured retrieval and answer generation over a small set of original passages.

The ablation experiments are run locally using Qwen3-32B on four
NVIDIA RTX A6000 GPUs. The main EnSI-RAG experiments use GPT-4.1-mini
and GPT-4.1 through the OpenAI API, for which provider-side hardware
details are unavailable.

\section{Conclusion}

We presented EnSI-RAG, an entity-structure-indexed retrieval-augmented generation framework for long-document question answering. Instead of indexing documents as arbitrary fixed-size chunks, EnSI-RAG constructs semantically contained passages and builds a query-independent index of the form $[\textit{entity}][\textit{entity\_type}][\textit{property} \mid \textit{relation} \mid \textit{aspect}: \textit{value}] \rightarrow \{\textit{psg\_id}\}$. This index provides structured access paths from future questions to sets of original supporting passages, improving evidence localization while preserving the LLM's role as the final semantic integrator. Across financial, legal, academic, and aggregation-heavy QA settings, EnSI-RAG is designed to provide a flexible middle ground between unstructured chunk-based RAG and fully structured database-style reasoning. More broadly, our results suggest that long-document QA benefits from separating evidence localization from answer synthesis: the index guides the system toward relevant passages, and the LLM integrates those passages into the final answer.

\bibliography{references}

\clearpage
\appendix

\renewcommand{\thetable}{A\arabic{table}}
\setcounter{table}{0}
\renewcommand{\thefigure}{A\arabic{figure}}
\setcounter{figure}{0}

\section{Appendix: Experimental Details}
\label{sec:supp-experimental-details}

\begin{table}[!h]
\centering
\normalsize
\setlength{\tabcolsep}{8pt}
\renewcommand{\arraystretch}{1.65}
\begin{tabular}{@{}p{0.16\textwidth}p{0.22\textwidth}p{0.57\textwidth}@{}}
\toprule
\textbf{Benchmark} & \textbf{Primary Focus} & \textbf{Description} \\
\midrule
Loong (Wang et al. 2024) &
Multi-document question answering &
A realistic long-context, multi-document question-answering benchmark spanning academic papers, legal documents, and financial reports. It includes four reasoning categories: Spotlight Locating, Comparison, Clustering, and Chain of Reasoning. These tasks require models to locate and integrate evidence across multiple related documents rather than retrieve an answer from a single context. \\
\midrule
Oolong (Bertsch et al. 2025) &
Long-context aggregation &
Evaluates long-context reasoning and information aggregation. Relevant information is distributed throughout long inputs, requiring a model to identify pertinent segments, reason locally over them, and aggregate the intermediate results into a final answer. The benchmark contains synthetic aggregation tasks (\textsc{Oolong-Synth}) and real-world aggregation tasks (\textsc{Oolong-Real}).
All Oolong results reported in the main paper are evaluated exclusively
on Oolong-Synth; we do not evaluate on Oolong-Real.\\
\bottomrule
\end{tabular}
\caption{Detailed descriptions of the benchmarks used in our experiments. Evaluated subset sizes and task types are reported in the main paper.}
\label{tab:supp-benchmark-details}
\end{table}

\clearpage

\begin{table}[!t]
\centering
\normalsize
\setlength{\tabcolsep}{8pt}
\renewcommand{\arraystretch}{1.80}
\begin{tabular}{@{}p{0.16\textwidth}p{0.25\textwidth}p{0.54\textwidth}@{}}
\toprule
\textbf{Benchmark} & \textbf{Evaluator or Metric} & \textbf{Protocol} \\
\midrule
Loong &
Official benchmark evaluation framework &
We use the official Loong evaluation framework, which is also adopted by SLIDERS, and report the average accuracy across the evaluated questions. \\
\midrule
Oolong &
LLM-as-a-judge and official deviation-based metric &
We follow the official Oolong protocol. Non-numeric questions are evaluated using an LLM-as-a-judge, whereas numeric aggregation questions use Oolong's official deviation-based metric, which assigns higher scores to predictions with smaller numerical errors. \\
\bottomrule
\end{tabular}
\caption{Evaluation protocols used for the two benchmarks. We follow each benchmark's official protocol.}
\label{tab:supp-evaluation-details}
\end{table}

\clearpage

\begin{table}[!t]
\centering
\small
\setlength{\tabcolsep}{8pt}
\renewcommand{\arraystretch}{1.42}
\begin{tabular}{@{}p{0.15\textwidth}p{0.22\textwidth}p{0.58\textwidth}@{}}
\toprule
\textbf{Baseline} & \textbf{Category} & \textbf{Description} \\
\midrule
RAG (Lewis et al. 2020) &
Conventional retrieval &
Retrieves semantically similar fixed-length document chunks and provides them to an LLM for answer generation. It represents the standard retrieval-based baseline. Because its retrieval units are determined by fixed chunk boundaries, related evidence may be separated and unrelated topics may be placed in the same chunk. \\
\midrule
LongRAG (Jiang, Ma, and Chen 2024) &
Larger-granularity retrieval &
Extends conventional RAG by retrieving substantially larger textual units. The increased retrieval granularity can reduce semantic fragmentation and preserve more context, but may also introduce more irrelevant information into the generation context. \\
\midrule
GraphRAG (Edge et al. 2024) &
Graph-guided retrieval &
Constructs a graph representation of the document collection in which entities and their relations form the retrieval structure. It performs graph-guided retrieval over connected entities before generating the final answer. \\
\midrule
BaseModel &
Direct long-context inference &
Directly performs long-context inference without external retrieval or preprocessing. The entire available document context is provided to the LLM, allowing us to compare structured retrieval and indexing with simply increasing the context supplied to the model. \\
\midrule
DocETL (Shankar et al. 2024) &
Document transformation &
Preprocesses long documents through LLM-driven extraction and transformation operations, converting unstructured documents into structured intermediate representations before question answering. Unlike retrieval-based methods, it focuses on making the document collection more query-friendly during preprocessing. \\
\midrule
Chain-of-Agent (Zhang et al. 2024) &
Multi-agent reasoning &
Decomposes long-document understanding among multiple coordinated LLM agents. Individual agents process different document portions or intermediate subtasks, and their outputs are progressively combined to produce the final answer. This design reduces the amount of context handled in any single model invocation. \\
\midrule
RLM (Zhang, Kraska, and Khattab 2025) &
Recursive reasoning &
Treats a long prompt as an external environment that the model can programmatically inspect and decompose. It recursively invokes the language model over selected parts of the prompt and aggregates the resulting intermediate outputs. \\
\midrule
SLIDERS (Joshi et al. 2026) &
Structured retrieval and reasoning &
The most closely related baseline and our primary comparison. It transforms long documents into reconciled relational tables through structured information extraction and cross-document reconciliation, then performs retrieval and reasoning over the resulting structured database. In contrast, EnSI-RAG uses structured records only as retrieval handles and generates answers from the retrieved original passages. \\
\bottomrule
\end{tabular}
\caption{Detailed descriptions of the baselines used in our experiments. The selected methods cover conventional retrieval-augmented generation, long-context inference, document transformation, agent-based reasoning, recursive reasoning, and structured retrieval.}
\label{tab:supp-baseline-details}
\end{table}

\clearpage

\onecolumn
\newcommand{\suppcorrect}{%
  \colorbox{green!15}{\textcolor{green!40!black}{\textbf{Correct}}}}
\newcommand{\suppincorrect}{%
  \colorbox{red!12}{\textcolor{red!55!black}{\textbf{Incorrect}}}}

\noindent\begin{minipage}{\textwidth}
\subsection{Case Studies}
\label{sec:supp-case-studies}

We present two representative cases in which the structure preserved during preprocessing determines whether the system can recover the complete answer. Table~\ref{tab:case-legal-full} examines legal judgment-result matching, and Table~\ref{tab:case-paper-full} examines citation-chain construction.

\begin{center}
\captionof{table}{Full case study on legal judgment-result matching. The task requires preserving benchmark-facing document identifiers in addition to legal-document attributes.}
\label{tab:case-legal-full}
\vspace{2pt}
\begingroup
\centering
\fontsize{8.3}{9.8}\selectfont
\setlength{\tabcolsep}{6pt}
\renewcommand{\arraystretch}{1.20}
\begin{tabular}{@{}p{0.21\textwidth}p{0.75\textwidth}@{}}
\toprule
\textbf{Case component} & \textbf{Details} \\
\midrule
\textbf{Question} &
The question provides 15 candidate judgment-result options and asks the system to assign each of 12 legal documents to the correct option. The answer must be a JSON mapping from benchmark document identifiers (e.g., \texttt{doc1}) to option identifiers (e.g., \texttt{doc8}). \\
\midrule
\textbf{Required operation} &
The task is not legal-case summarization or majority-result prediction. It requires the one-to-one mapping
\[
\textit{benchmark document label}\ \longrightarrow\ \textit{judgment-result option ID}.
\]
The answer key must use the benchmark-local document label rather than the full case title, and the value must be the option identifier rather than the judgment text. \\
\midrule
\textbf{SLIDERS representation} &
SLIDERS induces and reconciles a relational table organized around legal attributes, including case title, case number, court, legal category, and judgment result. This representation captures useful legal information but does not preserve the benchmark document label as the task-facing key. \\
\midrule
\textbf{SLIDERS behavior} &
The system produces related outputs such as judgment summaries, frequent-result predictions, or title-level mappings. These outputs may describe the cases correctly, but they do not provide the required \texttt{Doc-i} to \texttt{Result-j} mapping. \quad \suppincorrect \\
\midrule
\textbf{Why SLIDERS fails} &
The failure is caused by a representation mismatch. Identifiers such as \texttt{doc1} and \texttt{doc2} are benchmark-local labels rather than legal case titles. A table keyed primarily by case titles and legal attributes cannot directly recover those labels during generation, even when it contains the corresponding judgment text. \\
\midrule
\textbf{EnSI-RAG representation} &
Each legal document is represented by a metadata-centered passage that jointly preserves the benchmark document label, full case title, legal type, case category, court, case number, and judgment result. The resulting record makes the decision variable directly addressable:
\[
[\textit{document title}][\textit{legal type}]
[\textit{result}:\textit{judgment text}]
\rightarrow \{\textit{psg\_id}\}.
\]
The original passage also retains the benchmark label required for output formatting. \\
\midrule
\textbf{EnSI-RAG retrieval} &
At query time, EnSI-RAG recognizes judgment-result matching as the required task, parses the 15 candidate options, retrieves the judgment-result property for each document, and aligns it with the candidate options. Because both the canonical legal identity and benchmark-facing label are retained, the system can directly construct the requested mapping. \\
\midrule
\textbf{EnSI-RAG response} &
EnSI-RAG returns the required 12-entry JSON mapping from benchmark
document labels to judgment-result option identifiers. All 12
assignments match the gold answer; representative entries include
\texttt{"doc1":"doc8"}, \texttt{"doc2":"doc2"}, and
\texttt{"doc12":"doc15"}. \quad \suppcorrect \\
\midrule
\textbf{Ground truth} &
The benchmark's complete 12-entry document-to-option mapping. The
representative assignments above are drawn directly from this gold
mapping. \\
\midrule
\textbf{Key difference} &
SLIDERS organizes documents around induced legal attributes, whereas EnSI-RAG preserves the benchmark document identifier together with the canonical title, judgment-result property, and original evidence passage. The case is solved by retaining the correct retrieval and output handles before generation, rather than by adding more complex reasoning at generation time. \\
\bottomrule
\end{tabular}
\endgroup
\end{center}
\end{minipage}

\clearpage

\begin{table}[!t]
\centering
\small
\setlength{\tabcolsep}{8pt}
\renewcommand{\arraystretch}{1.36}
\caption{Full case study on paper citation-chain construction. The task requires a complete directed citation graph and canonical paper titles.}
\label{tab:case-paper-full}
\begin{tabular}{@{}p{0.21\textwidth}p{0.75\textwidth}@{}}
\toprule
\textbf{Case component} & \textbf{Details} \\
\midrule
\textbf{Question} &
Find the longest linear citation chain among five supplied papers and return the canonical paper titles in order. Only citation relations among the supplied papers should be considered, and non-linear branches should be ignored. \\
\midrule
\textbf{Required operation} &
The answer is not an unordered set of citation pairs. It requires constructing a directed graph over the five papers and selecting its longest valid linear path. The returned order runs from the earliest cited paper to the latest citing paper. \\
\midrule
\textbf{Gold chain} &
\textit{Towards Understanding Chain-of-Thought Prompting: An Empirical Study of What Matters}
$\rightarrow$
\textit{Large Language Models Encode Clinical Knowledge}
$\rightarrow$
\textit{Capabilities of GPT-4 on Medical Challenge Problems}
$\rightarrow$
\textit{Sparks of Artificial General Intelligence: Early experiments with GPT-4}
$\rightarrow$
\textit{ChatGPT believes it is conscious}. \\
\midrule
\textbf{SLIDERS representation} &
SLIDERS induces separate paper and citation tables containing extracted paper identifiers, titles, and citation pairs. In this case, some titles are not canonicalized and the extracted citation table contains only three of the four edges required for the gold chain. \\
\midrule
\textbf{Missing evidence} &
The citation from \textit{Capabilities of GPT-4 on Medical Challenge Problems} to \textit{Large Language Models Encode Clinical Knowledge} is missing. Once this edge is absent, no downstream query over the reconciled table can reconstruct the complete five-paper path. \\
\midrule
\textbf{SLIDERS response} &
\textit{Towards Understanding Chain-of-Thought Prompting}
$\rightarrow$
\textit{Large Language Models Encode Clinical Knowledge}
$\rightarrow$
\textit{Sparks of Artificial General Intelligence}
$\rightarrow$
\textit{ChatGPT believes it is conscious}. \quad \suppincorrect \\
\midrule
\textbf{EnSI-RAG representation} &
Each paper is assigned a canonical title and aliases derived from arXiv identifiers and filename variants. Reference passages are used to construct explicit forward \textit{cites} records and inverse \textit{cited-by} records:
\[
[\textit{source paper}][\textit{paper}]
[\textit{relation}:\textit{cites}=\textit{target paper}]
\rightarrow\{\textit{psg\_id}\}.
\]
Each relation retains provenance to its original reference passage. \\
\midrule
\textbf{Retrieved relations} &
EnSI-RAG recovers the four consecutive citation edges required by the gold chain, together with one additional non-linear edge. It constructs the directed citation graph, ignores the non-linear branch as requested, and selects the longest linear path. \\
\midrule
\textbf{EnSI-RAG response} &
\textit{Towards Understanding Chain-of-Thought Prompting: An Empirical Study of What Matters}
$\rightarrow$
\textit{Large Language Models Encode Clinical Knowledge}
$\rightarrow$
\textit{Capabilities of GPT-4 on Medical Challenge Problems}
$\rightarrow$
\textit{Sparks of Artificial General Intelligence: Early experiments with GPT-4}
$\rightarrow$
\textit{ChatGPT believes it is conscious}. \quad \suppcorrect \\
\midrule
\textbf{Ground truth} &
The complete five-paper chain shown above. \\
\midrule
\textbf{Key difference} &
SLIDERS reasons over the citation pairs that survive table extraction and reconciliation; a missing edge therefore makes the full chain unrecoverable. EnSI-RAG treats \textit{cites} and \textit{cited-by} as first-class directed relations, preserves canonical paper identities and aliases, and links every edge to original evidence. This relation-complete representation supports exact graph traversal and output in canonical-title form. \\
\bottomrule
\end{tabular}
\end{table}

\section{Additional Reproducibility Details}
\label{sec:supp-additional-reproducibility}

\begin{table}[t]
\centering
\small
\setlength{\tabcolsep}{5pt}
\renewcommand{\arraystretch}{1.15}
\begin{tabular}{p{0.13\textwidth}p{0.17\textwidth}p{0.59\textwidth}}
\hline
\textbf{Benchmark} & \textbf{Seed} & \textbf{Selection and verification} \\
\hline
Loong
& 42
& Finite subsets are sampled by difficulty level using the target proportions
40/30/20/10 for Levels 1--4.  The selected level and original
\texttt{source\_row\_index} are stored for every question, allowing the exact
subset to be reconstructed. \\
\hline
Oolong
& Not applicable
& We first filter the \texttt{test} split to questions whose document length
satisfies \texttt{context\_len=262144}, and then select the first 192 eligible
questions in the public dataset order.  The experiment entry point rejects any
selection whose split, context length, ordering, or selected question count
differs from this rule. All Oolong results reported in the main paper are evaluated exclusively
on Oolong-Synth; we do not evaluate on Oolong-Real.\\
\hline
\end{tabular}
\caption{Question selection and randomness controls. Oolong uses a deterministic
dataset slice and therefore does not require a sampling seed.}
\label{tab:supp-selection-reproducibility}
\end{table}

\paragraph{Verification of the Oolong slice.}
The public SLIDERS configuration applies the same deterministic selection
rule: it filters the \texttt{test} split by the 262,144-token document-length
condition and takes the first 192 eligible questions in dataset order.  Our
entry point therefore fixes \texttt{split=test}, sets both the minimum and
maximum context length to 262,144, and applies \texttt{start=0} and
\texttt{limit=192} after filtering.  Here, 192 is the size of the evaluated
subset; it is not the total number of questions in Oolong or necessarily the
total number satisfying the length condition.  Before evaluation, the program
verifies that the selected manifest contains exactly 192 questions, that every
question belongs to the \texttt{test} split, and that every selected document
has the required length.  The evaluator reconstructs the complete question
string used by the public SLIDERS Oolong driver and is invoked with a strict
slice check; any mismatch terminates the run without reporting a score.  We
additionally retain \texttt{selected\_queries.jsonl} and release the selected
question identifiers.

\paragraph{API and local execution configuration.}
Loong uses Top-12 retrieval. For Oolong-Synth, records are selected
deterministically by context-window identifier and subsequently
aggregated using query-focused metadata; the Oolong-Synth pipeline
does not instantiate an embedding model.  LLM-based preprocessing uses
GPT-4.1-mini, while retrieval planning and final answer generation use
GPT-4.1.  Oolong's LLM-based evaluator also uses GPT-4.1.  All API calls use
temperature 0.  These models are accessed through the OpenAI API; the
provider's server-side GPU, CPU, memory, and serving configuration are not
exposed to API users and are therefore unavailable to us.  We report the API
model identifiers, decoding parameters, and prompt templates that we control.
The local \texttt{octen-embedding-8B} model is used for retrieval on Loong's Financial dataset;
For other datasets, we select records deterministically by
\texttt{context\_window\_id} and does not instantiate the embedding model.

\paragraph{Number of runs.}
Each reported EnSI-RAG result is obtained from one complete run of the
corresponding configuration.  We therefore do not report across-run standard
deviations or statistical significance tests.
\clearpage
\section{Prompts}
\label{sec:supp-prompts}

This appendix provides the complete prompts used in the Loong and Oolong
experiments. The text inside every box is reproduced verbatim from the
experiment scripts. Braced expressions denote runtime substitutions in the
corresponding templates. Deterministic parsing, index construction, filtering,
and score aggregation do not invoke an LLM and therefore have no prompt to
report.

\subsection{Loong Prompts}
\label{sec:supp-prompts-loong}

\subsubsection{Company-Name Alias Extraction}
This prompt identifies bilingual company-name variants so that financial
records referring to the same company can share a canonical retrieval handle.

\begin{EnSIPromptBox}{Company-Name Alias Extraction: System Prompt}{EnSIBlue}
\begin{EnSIPromptText}
\PromptLine{You are a precise bilingual financial-report company-name alias extractor.}
\PromptBlank
\PromptLine{Your job is to identify all English or Chinese names that refer to the reporting company.}
\PromptLine{This includes the exact current registrant name and old/former/predecessor names if the filing says the company changed its name.}
\PromptLine{Return JSON only. Do not include explanations or markdown.}

\end{EnSIPromptText}
\end{EnSIPromptBox}

\begin{EnSIPromptBox}{Company-Name Alias Extraction: User Prompt}{EnSIBlue}
\begin{EnSIPromptText}
\PromptLine{Extract the company/registrant names from the filing excerpt below.}
\PromptBlank
\PromptLine{Rules:}
\PromptLine{- Include the exact current registrant name.}
\PromptLine{- Include former names only if the excerpt explicitly says the company was previously incorporated as, formerly known as, or changed its name from/to that name.}
\PromptLine{- For Chinese reports, recognize fields such as \CnCompanyNames, \CnFormerNames\ and \CnNameChange.}
\PromptLine{- Do not include subsidiaries, auditors, creditors, customers, officers, locations, or stock tickers unless they are also company names of the registrant.}
\PromptLine{- Keep names as clean display names, e.g. "Arvana Inc." and "Turinco, Inc.".}
\PromptBlank
\PromptLine{Return only this JSON schema:}
\PromptLine{\{}
\PromptLine{\hspace*{1.10em}"company\_names": ["name 1", "name 2"]}
\PromptLine{\}}
\PromptBlank
\PromptLine{Filing excerpt:}
\PromptLine{\{context\}}

\end{EnSIPromptText}
\end{EnSIPromptBox}

\subsubsection{Financial-Table Location}
This prompt locates financial tables and preserves the local context needed to
construct semantically contained row-level passages.

\begin{EnSIPromptBox}{Financial-Table Location: System Prompt}{EnSITeal}
\begin{EnSIPromptText}
\PromptLine{You are a bilingual indexing system for Loong Financial RAG.}
\PromptBlank
\PromptLine{Your job is NOT to answer questions and NOT to extract numerical answer values.}
\PromptLine{Your job is to inspect an English or Chinese financial-report chunk and identify}
\PromptLine{financial-statement tables that should become retrieval passages.}
\PromptBlank
\PromptLine{Important:}
\PromptLine{- The document may be an SEC Form 10-K, a Chinese annual report, or a Chinese quarterly report.}
\PromptLine{- Line numbers are global document line numbers even when only a chunk is supplied.}
\PromptLine{- Select real financial statement/value tables, not a table of contents or prose.}
\PromptLine{- One table must become one passage.}
\PromptLine{- Return the exact start/end line numbers for each table in the full filing.}
\PromptLine{- Extract only retrieval metadata: table title, years shown in the table, and metric row labels.}
\PromptLine{- Do NOT output row values, dollar values, percentages, share counts, or final answers.}
\PromptBlank
\PromptLine{Return JSON only. Do not include explanations, markdown, or \textless{}think\textgreater{}.}

\end{EnSIPromptText}
\end{EnSIPromptBox}

\begin{EnSIPromptBox}{Financial-Table Location: User Prompt}{EnSITeal}
\begin{EnSIPromptText}
\PromptLine{Read the following line-numbered chunk of a financial report}
\PromptLine{and identify all financial-statement tables that should be stored as passages.}
\PromptBlank
\PromptLine{The chunk keeps global line numbers from the full document. Return those global line numbers.}
\PromptBlank
\PromptLine{Table selection rules:}
\PromptLine{- A passage should be exactly one financial statement table.}
\PromptLine{- Include main financial statement tables, such as Balance Sheets, Statements of Operations, Statements of Income, Statements of Cash Flows, Statements of Stockholders' Equity/Deficit, Statements of Comprehensive Income/Loss, and similar tables.}
\PromptLine{- Chinese equivalents include \CnStatementTypes\ and their parent-company versions.}
\PromptLine{- Include notes tables only when they are actual financial line-item/value tables that may answer a financial metric question.}
\PromptLine{- Do NOT return the table of contents.}
\PromptLine{- Do NOT return prose paragraphs, audit report paragraphs, signatures, exhibit lists, or pure section headers.}
\PromptLine{- Do NOT extract numerical values into the metadata.}
\PromptLine{- Only extract metric labels that are actual financial line items with values. Do not extract pure section headers unless the header itself has reported values.}
\PromptBlank
\PromptLine{Return only a valid JSON object with this schema:}
\PromptLine{\{}
\PromptLine{\hspace*{1.10em}"tables": [}
\PromptLine{\hspace*{2.20em}\{}
\PromptLine{\hspace*{3.30em}"title": "short table title",}
\PromptLine{\hspace*{3.30em}"start\_line": 123,}
\PromptLine{\hspace*{3.30em}"end\_line": 145,}
\PromptLine{\hspace*{3.30em}"years": [2024, 2023],}
\PromptLine{\hspace*{3.30em}"metrics": ["Cash and cash equivalents", "Total assets"]}
\PromptLine{\hspace*{2.20em}\}}
\PromptLine{\hspace*{1.10em}]}
\PromptLine{\}}
\PromptBlank
\PromptLine{Line-numbered financial-report chunk:}
\PromptLine{\{numbered\_text\}}

\end{EnSIPromptText}
\end{EnSIPromptBox}

\subsubsection{Reference-Section Location}
This prompt locates the reference section of an academic paper before citation
relations and their source passages are constructed.

\begin{EnSIPromptBox}{Reference-Section Location: System Prompt}{EnSIGreen}
\begin{EnSIPromptText}
\PromptLine{You identify bibliography/reference sections in parsed academic papers.}
\PromptLine{Return JSON only. Do not explain. Do not use markdown.}

\end{EnSIPromptText}
\end{EnSIPromptBox}

\begin{EnSIPromptBox}{Reference-Section Location: User Prompt}{EnSIGreen}
\begin{EnSIPromptText}
\PromptLine{Find the bibliography/references section in the following parsed academic paper tail.}
\PromptLine{Return ONLY a JSON object with this schema:}
\PromptLine{\{}
\PromptLine{\hspace*{1.10em}"found": true/false,}
\PromptLine{\hspace*{1.10em}"start\_quote": "an exact substring copied from the first 80-200 characters of the references section",}
\PromptLine{\hspace*{1.10em}"end\_quote": "an exact substring copied from the last 80-200 characters of the references section"}
\PromptLine{\}}
\PromptBlank
\PromptLine{Rules:}
\PromptLine{- The references section is the bibliography at the end of the paper, not citations in Related Work.}
\PromptLine{- Do not paraphrase. start\_quote and end\_quote must be exact substrings from the input.}
\PromptLine{- If there is no references/bibliography section, return \{"found": false, "start\_quote": "", "end\_quote": ""\}.}
\PromptBlank
\PromptLine{PAPER TAIL START}
\PromptLine{\{tail\}}
\PromptLine{PAPER TAIL END}

\end{EnSIPromptText}
\end{EnSIPromptBox}

\subsubsection{Retrieval Planning}
This prompt converts a Loong question into structured retrieval items aligned
with the entity-structure index.

\begin{EnSIPromptBox}{Retrieval Planning: System Prompt}{EnSIPurple}
\begin{EnSIPromptText}
\PromptLine{You are the single generic retrieval planner for EnSI-RAG.}
\PromptBlank
\PromptLine{The corpus has already been preprocessed into dataset-specific passages, but all records share one schema:}
\PromptLine{[entity][entity\_type][property|relation|aspect: value] -\textgreater{} passage\_id}
\PromptBlank
\PromptLine{Dataset-specific conventions:}
\PromptLine{- paper: entity\_type=paper. Important relations are cites and cited-by. References section is indexed as references; body is content.}
\PromptLine{- financial: entity\_type=financial\_statement\_year. Entity is year + company, e.g. "2024 Apple Inc.". The association is the metric/aspect only. Numeric values remain in the original row passage text.}
\PromptLine{- legal: entity is the legal document title. Entity type is legal\_type. Important properties are case, sub\_case, subsub\_case, combined\_case, court, number, result, metadata, and content.}
\PromptBlank
\PromptLine{The question may be Chinese. Understand it directly. For a Chinese financial}
\PromptLine{question over an English SEC filing, translate each requested metric into the}
\PromptLine{canonical English filing label in association\_name and add common English}
\PromptLine{alternatives in metric\_aliases. Do not put only Chinese text in association\_name.}
\PromptBlank
\PromptLine{Convert the question into retrieval items. Do not answer. Return JSON only.}

\end{EnSIPromptText}
\end{EnSIPromptBox}

\begin{EnSIPromptBox}{Retrieval Planning: User Prompt}{EnSIPurple}
\begin{EnSIPromptText}
\PromptLine{Dataset type: \{dataset\_type\}}
\PromptBlank
\PromptLine{Question:}
\PromptLine{\{question\}}
\PromptBlank
\PromptLine{Instruction:}
\PromptLine{\{instruction\}}
\PromptBlank
\PromptLine{Provided documents:}
\PromptLine{\{docs\}}
\PromptBlank
\PromptLine{Return only this JSON schema:}
\PromptLine{\{}
\PromptLine{\hspace*{1.10em}"retrieval\_items": [}
\PromptLine{\hspace*{2.20em}\{}
\PromptLine{\hspace*{3.30em}"entity": "entity name if known; for financial use '\textless{}year\textgreater{} \textless{}company\textgreater{}' when year and company are known",}
\PromptLine{\hspace*{3.30em}"entity\_type": "paper | financial\_statement\_year | legal\_type string | empty if unknown",}
\PromptLine{\hspace*{3.30em}"association\_kind": "property | relation | aspect | empty if unknown",}
\PromptLine{\hspace*{3.30em}"association\_name": "property/relation/aspect name such as cites, cited-by, content, case, combined\_case, net income",}
\PromptLine{\hspace*{3.30em}"metric\_aliases": ["English filing-label alternatives, especially when the financial question is Chinese"],}
\PromptLine{\hspace*{3.30em}"association\_value": "target value or label if known",}
\PromptLine{\hspace*{3.30em}"company": "financial company if relevant",}
\PromptLine{\hspace*{3.30em}"year": null,}
\PromptLine{\hspace*{3.30em}"target\_value": "numeric/legal/paper value constraint if any"}
\PromptLine{\hspace*{2.20em}\}}
\PromptLine{\hspace*{1.10em}],}
\PromptLine{\hspace*{1.10em}"operation": "lookup | classify | compare | trend | identify\_company | citation\_reference | citation\_chain | other",}
\PromptLine{\hspace*{1.10em}"answer\_requirements": "short description of the final output format"}
\PromptLine{\}}

\end{EnSIPromptText}
\end{EnSIPromptBox}

\subsubsection{Final-Answer Generation}
This prompt generates the final Loong response from the retrieval plan,
matched index records, and retrieved original passages.

\begin{EnSIPromptBox}{Final-Answer Generation: System Prompt}{EnSIBurgundy}
\begin{EnSIPromptText}
\PromptLine{You are the unified final-answer generator for EnSI-RAG.}
\PromptBlank
\PromptLine{You must answer using only the retrieved original evidence passages and the retrieved index records.}
\PromptLine{The records are retrieval hints; the original passage text is the evidence.}
\PromptLine{Do not use outside knowledge.}
\PromptLine{Preserve the answer format requested by the question/instruction.}
\PromptLine{When the expected answer is JSON, return only valid JSON with no markdown.}
\PromptLine{When the expected answer is a short text answer, return only the concise final answer.}
\PromptLine{Understand Chinese questions directly and answer in the language and exact}
\PromptLine{format requested by the original instruction. Financial evidence may use}
\PromptLine{English filing labels even when the question is Chinese.}
\PromptBlank
\PromptLine{Financial-classification rule:}
\PromptLine{- When the operation is classification over multiple provided companies, include every provided company exactly once.}
\PromptLine{- Do not silently omit a company. Follow the requested grouped JSON schema rather than returning a partial list.}
\PromptBlank
\PromptLine{Paper-output rule:}
\PromptLine{- For paper tasks, never output an arXiv id, filename, filename stem, or local path as the final paper answer.}
\PromptLine{- Always output the canonical full paper title from the provided/retrieved paper metadata.}
\PromptLine{- Preserve the benchmark title style, e.g. "\# TinyLlama: An Open-Source Small Language Model".}
\PromptBlank
\PromptLine{Legal-output rule:}
\PromptLine{- For Loong legal classification tasks, use benchmark document labels such as "\CnJudgmentLabelSix" unless the question explicitly asks for actual full titles.}
\PromptLine{- Do not output full legal document titles when the expected/gold format uses \CnJudgmentLabel\ labels.}

\end{EnSIPromptText}
\end{EnSIPromptBox}

\begin{EnSIPromptBox}{Final-Answer Generation: User Prompt}{EnSIBurgundy}
\begin{EnSIPromptText}
\PromptLine{Dataset type: \{dataset\_type\}}
\PromptBlank
\PromptLine{Original instruction:}
\PromptLine{\{instruction\}}
\PromptBlank
\PromptLine{Original question:}
\PromptLine{\{question\}}
\PromptBlank
\PromptLine{Provided documents:}
\PromptLine{\{docs\}}
\PromptBlank
\PromptLine{Retrieval plan:}
\PromptLine{\{plan\_json\}}
\PromptBlank
\PromptLine{Retrieved index records:}
\PromptLine{\{record\_evidence\}}
\PromptBlank
\PromptLine{Retrieved original passages:}
\PromptLine{\{passage\_evidence\}}
\PromptBlank
\PromptLine{Answer the original question based solely on the retrieved evidence.}

\end{EnSIPromptText}
\end{EnSIPromptBox}

\clearpage
\subsection{Oolong Prompts}
\label{sec:supp-prompts-oolong}

\subsubsection{Atomic-Instance Record Extraction}
This prompt converts Oolong atomic instances into structured records while
retaining their provenance and context-window identifiers.

\begin{EnSIPromptBox}{Atomic-Instance Record Extraction: System Prompt}{EnSIGreen}
\begin{EnSIPromptText}
\PromptLine{You are an exact text classification engine. Classify every supplied instance according to the task definition. Return JSON only. Never omit an item and never invent labels outside ALLOWED\_LABELS.}

\end{EnSIPromptText}
\end{EnSIPromptBox}

\begin{EnSIPromptBox}{Atomic-Instance Record Extraction: User Prompt}{EnSIGreen}
\begin{EnSIPromptText}
\PromptLine{TASK\_DEFINITION:}
\PromptLine{\{definition\}}
\PromptBlank
\PromptLine{ALLOWED\_LABELS:}
\PromptLine{\{json.dumps(labels, ensure\_ascii=False)\}}
\PromptBlank
\PromptLine{INSTANCES:}
\PromptLine{\{json.dumps(payload, ensure\_ascii=False)\}}
\PromptBlank
\PromptLine{Return exactly a JSON array of objects: [\{"id":"...","label":"one allowed label","confidence":0.0\}].}

\end{EnSIPromptText}
\end{EnSIPromptBox}

\subsubsection{Final-Answer Generation}
This prompt produces an Oolong answer from the retrieved records associated
with the selected context windows.

\begin{EnSIPromptBox}{Final-Answer Generation: System Prompt}{EnSIBurgundy}
\begin{EnSIPromptText}
\PromptLine{You are the final reasoning and answer-generation component of an Entity-Structure Indexed RAG system.}
\PromptBlank
\PromptLine{You MUST answer the original question using only the retrieved EnSI metadata supplied by the user.}
\PromptLine{The metadata was computed from LLM-extracted index entries. Each index entry has the form:}
\PromptLine{[entity][oolong\_instance][label: value]}
\PromptLine{and may also carry user and date metadata.}
\PromptBlank
\PromptLine{Important rules:}
\PromptLine{1. Do not use outside knowledge.}
\PromptLine{2. Do not ask for the original long context.}
\PromptLine{3. Perform all comparisons, ranking, arithmetic, and final decision yourself from the supplied metadata.}
\PromptLine{4. Do not output explanations, reasoning, markdown, or code fences.}
\PromptLine{5. Return exactly one answer line in the requested format.}
\PromptLine{6. Never invent a label, user, date, or number that is absent from the supplied metadata.}

\end{EnSIPromptText}
\end{EnSIPromptBox}

\begin{EnSIPromptBox}{Final-Answer Generation: User Prompt}{EnSIBurgundy}
\begin{EnSIPromptText}
\PromptLine{ORIGINAL QUESTION:}
\PromptLine{\{str(row.get('question', ''))\}}
\PromptBlank
\PromptLine{TASK METADATA:}
\PromptLine{\{json.dumps(\{'dataset': row.get('dataset'), 'task\_group': row.get('task\_group'), 'task': row.get('task'), 'answer\_type': row.get('answer\_type')\}, ensure\_ascii=False, indent=2)\}}
\PromptBlank
\PromptLine{RETRIEVED ENSI METADATA:}
\PromptLine{\{json.dumps(metadata, ensure\_ascii=False, indent=2)\}}
\PromptBlank
\PromptLine{OUTPUT FORMAT:}
\PromptLine{\{\_answer\_format\_instruction(row.get('answer\_type'))\}}
\PromptLine{Return only that one line.}

\end{EnSIPromptText}
\end{EnSIPromptBox}

\subsubsection{Soft Evaluation}
This prompt implements the LLM-based evaluator used for non-numeric Oolong
questions.

\begin{EnSIPromptBox}{Soft Evaluation: System Prompt}{EnSIOrange}
\begin{EnSIPromptText}
\PromptLine{Evaluate the predicted answer against the gold answer. The predicted answer should match the gold answer.}

\end{EnSIPromptText}
\end{EnSIPromptBox}

\begin{EnSIPromptBox}{Soft Evaluation: User Prompt}{EnSIOrange}
\begin{EnSIPromptText}
\PromptLine{\# Question}
\PromptLine{\{question\}}
\PromptBlank
\PromptLine{\# Gold Answer}
\PromptLine{\{\_json\_text(gold\_answer)\}}
\PromptBlank
\PromptLine{\# Predicted Answer}
\PromptLine{\{predicted\_answer\}}

\end{EnSIPromptText}
\end{EnSIPromptBox}

\subsubsection{Numeric-Answer Extraction}
This prompt extracts a numeric prediction before the official
deviation-based Oolong metric is applied.

\begin{EnSIPromptBox}{Numeric-Answer Extraction: System Prompt}{EnSIBlue}
\begin{EnSIPromptText}
\PromptLine{Given the question and the assistant's full answer, extract only the final numeric answer. Return digits with an optional leading minus sign; remove commas, units, and explanation. If no numeric answer is present, return null.}

\end{EnSIPromptText}
\end{EnSIPromptBox}

\begin{EnSIPromptBox}{Numeric-Answer Extraction: User Prompt}{EnSIBlue}
\begin{EnSIPromptText}
\PromptLine{\# Question}
\PromptLine{\{question\}}
\PromptBlank
\PromptLine{\# Predicted Answer}
\PromptLine{\{predicted\_answer\}}

\end{EnSIPromptText}
\end{EnSIPromptBox}

\end{document}